\documentclass[runningheads]{llncs}
\usepackage[T1]{fontenc}
\usepackage{graphicx}
\usepackage{xspace}
\usepackage{amsmath}
\usepackage{subcaption}

\usepackage{caption}
\usepackage{tabularx}
\usepackage{multirow}
\usepackage{booktabs}
\usepackage{setspace}
\usepackage{makecell}
\usepackage{dblfloatfix}
\usepackage[utf8]{inputenc}
\usepackage[font=small,labelfont=bf]{caption}
\newcommand{\ie}{i.\,e.,\xspace}
\newcommand{\eg}{e.\,g.,\xspace}

\begin{document}
\title{Memorization of Named Entities in \\ Fine-tuned BERT Models}
\titlerunning{Memorization of Named Entities in BERT Models}
%
\author{Andor Diera \inst{1}\orcidID{0009-0001-3959-493X} \and
Nicolas Lell \inst{1}\orcidID{0000-0002-6079-6480} \and
Aygul Garifullina \inst{2}\orcidID{0009-0005-0160-5479}\and Ansgar Scherp \inst{1}\orcidID{0000-0002-2653-9245}}
\authorrunning{Diera et al.}
%
\institute{Ulm University, Ulm, Germany \email{{firstname.lastname@uni-ulm.de}} \and
BT, Ipswich, United Kingdom
\email{aygul.garifullina@bt.com}}
\maketitle              
\begin{abstract}
Privacy preserving deep learning is an emerging field in machine learning that aims to mitigate the privacy risks in the use of deep neural networks. One such risk is \textit{training data extraction} from language models that have been trained on datasets, which contain personal and privacy sensitive information. In our study, we investigate the extent of named entity memorization in fine-tuned BERT models. We use single-label text classification as representative downstream task and employ three different fine-tuning setups in our experiments, including one with Differential Privacy (DP). We create a large number of text samples from the fine-tuned BERT models utilizing a custom sequential sampling strategy with two prompting strategies. We search in these samples for named entities and check if they are also present in the fine-tuning datasets. We experiment with two benchmark datasets in the domains of emails and blogs. We show that the application of DP has a detrimental effect on the text generation capabilities of BERT.
Furthermore, we show that a fine-tuned BERT does not generate more named entities specific to the fine-tuning dataset than a BERT model that is pre-trained only. This suggests that BERT is unlikely to emit personal or privacy sensitive named entities. Overall, our results are important to understand to what extent BERT-based services are prone to training data extraction attacks\footnote{Source code and datasets are available at: \url{https://github.com/drndr/bert\_ent\_attack}}. 

\keywords{language models \and training data extraction \and privacy preserving deep learning}
\end{abstract}
\section{Introduction}
\label{sec:introduction}
Deep Neural Networks (DNNs) became the \textit{de facto} tool for achieving state-of-the-art performance in many research domains such as computer vision and natural language processing (NLP).
Although utilizing large volumes of training data is one of the main driving factors behind the great performance of DNNs, publishing these models to the public raises some serious privacy concerns regarding private and confidential information present in the training data~\cite{liu2021machine}. These privacy concerns are especially relevant for large Language Models (LMs) which form the basis of state-of-the-art technologies in many NLP tasks.
 Recent versions of these models are usually first pre-trained in a task-agnostic self-supervised manner. The latest large LMs use a corpus size ranging from hundreds of gigabytes to several terabytes of text~\cite{brown2020language,raffel2019exploring} during this self-supervised process. The sheer size of these datasets makes it near impossible for researchers to remove all confidential information which may be present in the corpus. A recent study has shown that it is possible to extract personal information from some large LMs, even if that given information has only appeared once in the training corpus~\cite{carlini2021extracting}.
While the training cost of these large LMs became so prohibitively expensive that only the biggest tech companies can afford it~\cite{sharir2020cost}, pre-trained LMs are commonly used in businesses that work with huge amounts of text data. These businesses include banks, telecommunications, and insurance companies, which often handle a great amount of personal and privacy sensitive data. In practice, pre-trained LMs are fine-tuned on a business-specific dataset using some downstream task (such as text-classification, question-answering, or natural language inference) before deployment~\cite{devlin2018bert}. 
Although the fine-tuning may mitigate some of the unintended memorization of the original dataset used in pre-training, it raises new concerns regarding the personal and privacy sensitive information in the business-specific dataset used for the fine-tuning process~\cite{carlini2021extracting}.
%
Privacy Preserving Deep Learning (PPDL) is a common term used for methods aiming to mitigate general privacy concerns present in the use of DNNs. Multiple approaches have been proposed to achieve PPDL~\cite{mireshghallah2020privacy}, but there is no perfect solution to this problem, with each method having its own challenges and limitations. The most popular techniques include Federated Learning~\cite{mcmahan2016federated}, the application of Differential Privacy (DP)~\cite{abadi2016deep}, 
encryption~\cite{gilad2016cryptonets}, 
and data anonymization~\cite{sweeney2002k}. 

We investigate whether it is possible to extract personal information from one of the most popular modern LMs, BERT~\cite{devlin2018bert}. BERT is an auto-encoder transformer that is mostly used for natural language understanding tasks. Since BERT is less adept at generating long, coherent text sequences~\cite{wang2019bert}, we focus our study on the generation and extraction of named entities.
We conduct our experiments on three typical fine-tuning setups to understand the privacy risks involved in using BERT for commercial purposes.
We consider 
fine-tuning all layers of BERT (\textit{Full}), 
fine-tuning only the last encoding layer and the classifier head of BERT (\textit{Partial}), and
partial fine-tuning but with a privacy preserving optimizer (\textit{Differentially Private}, short: \textit{DP}).
As a privacy preserving optimizer, we employ the established differentially private stochastic gradient descent (DPSGD) algorithm~\cite{abadi2016deep}, which we discuss in detail in Section~\ref{sec:dpsgd}.
We compare the fine-tuning setups with a pre-trained only BERT base model.
We experiment with two benchmark datasets in the domains of emails (Enron Email corpus~\cite{klimt2004introducing}) and blogs (Blog Authorship Corpus~\cite{schler2006effects}).
For triggering entity extraction, we use two prompting techniques.
The \textit{naive prompting} is based on randomly selected text from the web, while the \textit{informed prompting} uses actual text from the datasets' test sets.
Each experimental setup is assessed with regard to its performance on the down-stream task, \ie single-label text classification, and the extent of named entity memorization.
In summary, our experiments show:

\begin{itemize}

\item The memorization rate of named entities in the fine-tuned BERT models is less than 10\% in both datasets across all setups.
Interestingly, the fine-tuned models do not emit more entities from the fine-tuning datasets than a pre-trained only BERT model.

\item When comparing the informed prompting versus the naive prompting, the BERT models consistently generate more named entities when using naive prompts.
Thus a potential attacker does not require prior knowledge about the training dataset of a model. 

\item Applying differentially private fine-tuning results in a strong drop in the amount of memorized entities at the cost of downstream task performance.
It effectively reduces the amount of entity memorization in fine-tuned BERT models.
\end{itemize}

Below, we discuss the related work. 
Our framework and methods for extracting named entities from fine-tuned BERT models are described in Section~\ref{sec:methods}. Section~\ref{sec:experiments} introduces the datasets and the details of the experiments. The results are described in Section~\ref{sec:results} and discussed in Section~\ref{sec:discussion}.

\section{Related Work}
\label{sec:related_work}

\subsection{Language Models and Text Generation}
Modern large LMs rely on two core concepts that led to their dominance in the NLP field: the focus on the self-attention mechanism in the DNN architecture and the introduction of large-scale task-agnostic pre-training to learn general language representations~\cite{wolf2020transformers}.
Self-attention is used for modeling dependencies between different parts of a sequence. A landmark study in 2017~\cite{vaswani2017attention} has shown that self-attention was the single most important part of the state-of-the-art NLP models of that time. 
It introduced a new family of models called transformers, which rely solely on stacked layers of self-attention and feed-forward layers. Besides the state-of-the-art performances, another great advantage of the transformer architecture is that unlike a recurrent architecture, it allows for training parallelization. The ability to parallelize training, alongside the significant increase in computational power allowed these models to train on larger datasets than once was possible. Since supervised training requires labeled data, self-supervised pre-training with supervised fine-tuning became the standard approach when using these models~\cite{mao2020survey}.

State-of-the-art transformers can be divided into three main categories based on their pre-training approach \cite{zhang2022survey}. 
Auto-re\-gres\-sive models use the classical language modeling pre-training task of next word prediction. Auto-encoding models are pre-trained by reconstructing sequences that have been corrupted in some way. 
Sequence-to-sequence models usually employ objectives of encoding-decoding models for pre-training, like replacing random sequences in a text with one special token with the objective of predicting that given sequence~\cite{raffel2019exploring}.

\subsubsection{BERT}
A major limitation of the auto-regressive models is that during pre-training they learn a unidirectional language model. 
In these models, tokens are restricted to only attend to other tokens left to them.
In contrast, BERT is an auto-encoder transformer model that uses an attention mechanism on the entire input sequence~\cite{devlin2018bert}.
%
This model utilizes Masked Language Modeling (MLM) and Next Sentence Prediction (NSP) as pre-training tasks. 
In MLM, some tokens are randomly removed from the input sequence and the model is trained to predict the removed tokens using context from both directions. 

Although the parameter count of BERT is greatly surpassed by more recent large LMs (such as the new GPT models \cite{brown2020language,ouyang2022training}), BERT is still one of the most common baselines in many NLP benchmark tasks. The strong performance coupled with the fact that the model is democratized and has publicly available pre-trained implementations makes BERT a popular choice of NLP model both in industry and academia. Since its original release, there have been dozens of follow-up studies and models published~\cite{rogers2020primer}.
The most notable variants include RoBERTa \cite{liu2019roberta}, DistilBERT \cite{sanh2019distilbert}, DeBERTa \cite{he2020deberta}, and domain-specific models such as SciBERT \cite{beltagy2019scibert} or ClinicalBERT \cite{alsentzer2019publicly}.

\subsubsection{Natural Language Generation}
\label{sec:nlg}
Natural Language Generation (NLG) is a subfield of NLP that is focused on producing natural language text that enables computers to write like humans~\cite{zhang2022survey}. 
Although auto-regressive transformer models are the standard choice for the task of NLG (since they are already trained to predict the next token based solely on previous tokens in a sequence), it has been shown that BERT can also be utilized to generate reasonably coherent text. Wang and Cho~\cite{wang2019bert} designed a generation strategy for BERT based on Gibbs sampling~\cite{geman1984stochastic}, where given a seed sequence, tokens at random positions are masked and replaced by new tokens based on the sampling technique. 
Another generation strategy developed for auto-encoding transformers~\cite{ghazvininejad2019mask} is to use a fully masked sequence as input and predict all tokens at once. 
Subsequently, tokens with the lowest probability are iteratively re-masked and replaced with a newly computed token.

\subsection{Privacy Attacks in Machine Learning}
Privacy attacks in machine learning denote a specific type of adversarial attack, which aim to extract information from a trained model. Based on recent surveys in the field~\cite{de2020overview,liu2021machine,rigaki2020survey}, these attacks can be divided into five main categories.

\subsubsection{Training Data Extraction Attacks}
Training data extraction attacks aim to reconstruct training datapoints, but unlike model inversion attacks, the goal is to retrieve verbatim training examples and not just ``fuzzy'' class representatives~\cite{carlini2021extracting}. 
These attacks are best suited for generative sequence models such as LMs. Initially these attacks have been designed for small LMs using academic datasets~\cite{carlini2019secret,zanella2020analyzing,thakkar2020understanding}. The aim of these studies was to measure the presence of specific training datapoints in the text samples generated by the models.
A common approach to measure the extent of this unintended memorization is to insert so-called ``canaries'' (artificial datapoints) into the training datasets and quantify their occurrence during sequence completion~\cite{carlini2019secret}.
Since these initial studies were based on smaller models trained with a high number of epochs, it was assumed that this kind of privacy leakage must be correlated with overfitting~\cite{zanella2020analyzing}. 
However, a follow-up study using the GPT-2 model, which is trained on a very large corpus for only a few epochs, showed that even state-of-the-art large LMs are susceptible to these kinds of attacks. 
Using the pre-trained GPT-2 model, Carlini et al.~\cite{carlini2021extracting} were able to generate and select sequence samples which contained low \textit{$k$-eidetic} data-points (data points that occur $k$ times in the training corpus). 
A study by Lehman et al.~\cite{lehman2021does} on Clinical BERT attempted to extract patient-condition association using both domain-specific template infilling and the text generation methods inspired by the text extraction research done on GPT-2~\cite{carlini2021extracting} and the BERT specific text generation technique proposed by Wang and Cho~\cite{wang2019bert}. 
Their methods were not successful in reliably extracting privacy sensitive information (patient-condition associations) from Clinical BERT, but it remains inconclusive whether it is due to the limitations in their method or in the linguistic capabilities of BERT.

\subsubsection{Membership Inference Attacks}
The goal of a membership inference attack is to determine whether or not an individual data instance is part of the training dataset for a given model. This attack typically assumes a black-box query access to the model. 
The common approach to this type of attack is to use a shadow training technique to imitate the behavior of a specific target model. 
In shadow training, a model (shadow model) is trained on a dataset that has a disjoint but identically formatted training data as the target model.
The trained inference model is then used to recognize differences on the target model predictions between inputs used for training and inputs not present in the training data~\cite{shokri2017membership}.

\subsubsection{Model Extraction Attacks}
The adversarial aim of a model extraction attack is to duplicate (\ie ``steal'') a given machine learning model.
It achieves this by training a function f' that is approximating the function f of the attacked model~\cite{liu2021machine}. 
A shadow training scheme has been shown to successfully extract popular machine learning models such as logistic regression, decision trees, and neural networks, using only black-box query access~\cite{tramer2016stealing}. Other works have proposed methods to extract information about hyperparameters~\cite{wang2018stealing} and properties of the architecture~\cite{oh2019towards} in neural networks.

\subsubsection{Model Inversion Attacks}
The idea behind model inversion attacks is that an adversary can infer sensitive information about the input data using a target model's output. These attacks can be used to extract input features and/or reconstruct prototypes of a class (in case the inferred feature characterize an entire class), given a white-box access to the model and knowledge about the target labels with some auxiliary information of the training data\cite{fredrikson2015model}.

\subsubsection{Property Inference Attacks}
The goal of property inference attacks is to infer some hidden property of a training dataset that the owner of the target model does not intend to share (such as feature distribution or training bias). Initially, property inference attacks were applied on discriminative models with white-box access \cite{parisot2021property}. A more recent work has extended the method to work on generative models with black-box access~\cite{parisot2021property}.

\subsection{Privacy Preserving Deep Learning}Based on the literature~\cite{de2020overview,liu2021machine,rigaki2020survey}, PPDL methods can be divided into four main categories.


\subsubsection{Differentially Private Learning}
Differential Privacy (DP) is a rigorous mathematical definition of privacy in the context of statistical and machine learning analysis. 
It addresses the challenge of ``learning nothing about an individual while learning useful information about the population''~\cite{dwork2014algorithmic}. 
In machine learning, DP algorithms aim to obfuscate either the training data~\cite{zhang2018privacy} or the model~\cite{rubinstein2009learning} by adding noise. 
Since directly adding noise to DNN parameters may significantly harm its utility, the best and most common place for applying DP in deep learning is the gradients \cite{zhu2020more}.
Abadi et al.~\cite{abadi2016deep} proposed an efficient training algorithm with a modest privacy budget called Differentially Private Stochastic Gradient Descent (DPSGD). 
DPSGD ensures DP by cutting the gradients to a maximum L2 norm for each layer and then adding noise to the gradients. 
Although DPSGD comes with increased computational cost and performance loss, variations of this algorithm~\cite{dupuy2021efficient,davody2020robust} still belong to the cutting-edge of PPDL research.

\subsubsection{Encryption}
Cryptography-based methods can be divided into two subcategories, depending whether the target of the encryption is the training data~\cite{gilad2016cryptonets} or the model~\cite{aono2017privacy}. Regardless of the target, most existing approaches use homomorphic encryption, which is a special kind of encryption scheme that allows computations to be performed on encrypted data without decrypting it in advance~\cite{acar2018survey}. Since training a DNN is already computationally expensive, adding homomorphic encryption to the process raises major challenges as it increases training times by at least an order of magnitude~\cite{liu2021machine}.

\subsubsection{Data Anonymization}
Data Anonymization techniques aim to remove all Personally Identifiable Information (PII) from a dataset. The common approach to achieve this is to remove attributes that are identifiers and mask quasi-identifier attributes~\cite{xu2014survey}. The popular $k$-anonymity algorithm~\cite{sweeney2002k} works by suppressing identifiers (\ie replacing them with an asterisk) and generalizing quasi-identifiers with a broader category which has a frequency of at least $k$ in the dataset. 
Although data anonymization techniques were developed for structured data, it is possible to adapt them to unstructured text data~\cite{hassan2018anonymization} as well as jointly anonymizing structured data and unstructured text data~\cite{DBLP:conf/doceng/SinghoferGKS21}.

\subsubsection{Aggregation}
Aggregation methods are generally used along with distributed
learning, in which multiple parties train on the same  machine learning task while aiming to keep their respective datasets private~\cite{liu2021machine}. Although aggregation methods can provide data security during distributed training, their privacy preserving aspects are more limited than other PPDL approaches.

\section{Extracting Named Entities From BERT}
\label{sec:methods}
\begin{figure}[!th]
    \centering
    \includegraphics[width=0.85\linewidth]{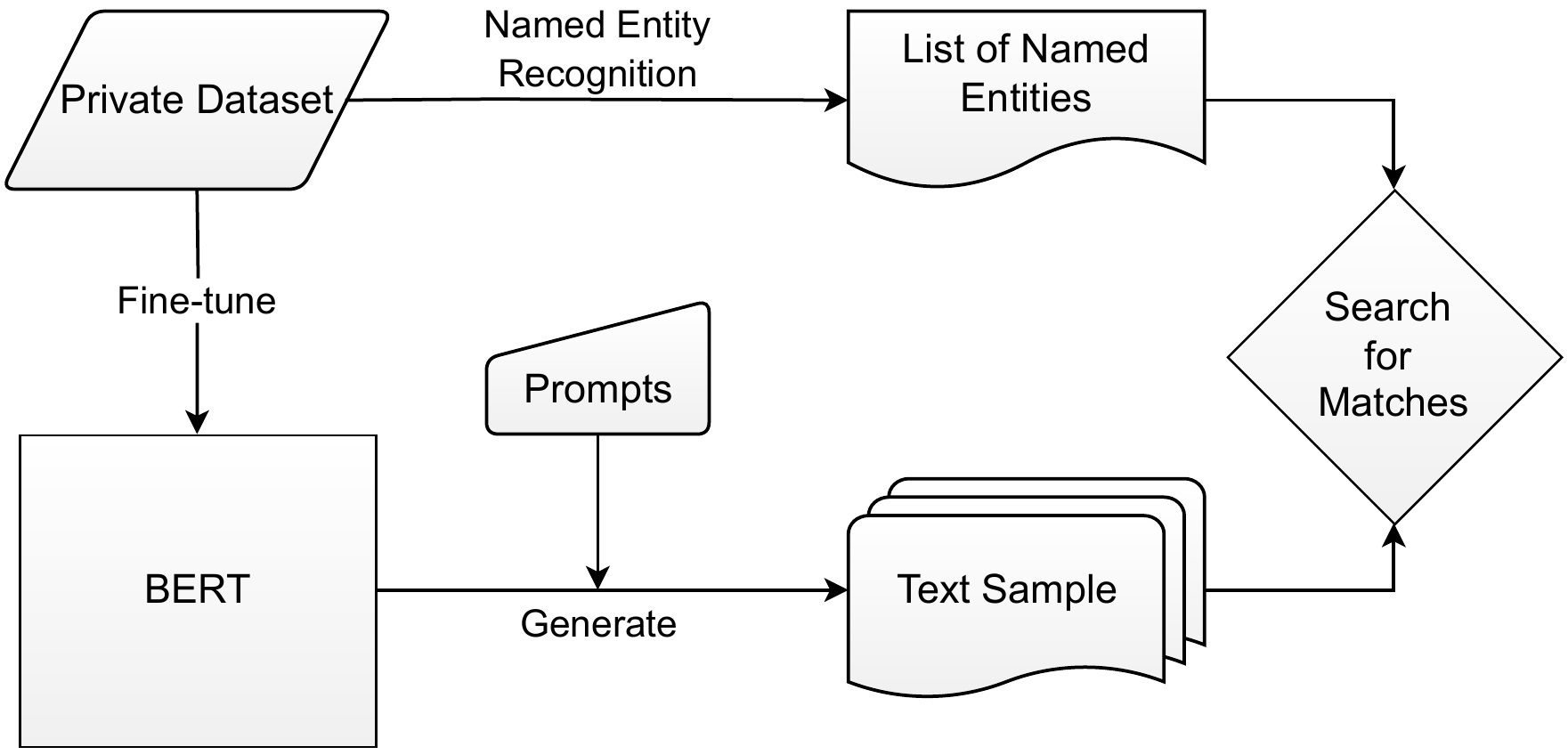}
    \caption[Pipeline for extracting named entities]{An illustration of our framework for extracting training data entities from BERT. 
    First, we fine-tune a pre-trained BERT on a private dataset. 
    Next, we generate text samples from the fine-tuned model using prompts. 
    Finally, we search the generated samples for the named entities that occur in the private dataset.}
    \label{fig:pipeline}
\end{figure}

In order to extract the named entities of the fine-tuning dataset from the BERT model, we present the experimental pipeline depicted in Figure~\ref{fig:pipeline}.
The pipeline consists of three phases:  
fine-tuning (including a privacy preserving approach using Differential Privacy), 
text generation from the fine-tuned models, and evaluation of the named entity memorization.

\subsection{Fine-tuning}
\begin{figure}[!th]
    \centering
    \includegraphics[width=1\linewidth, height=8cm]{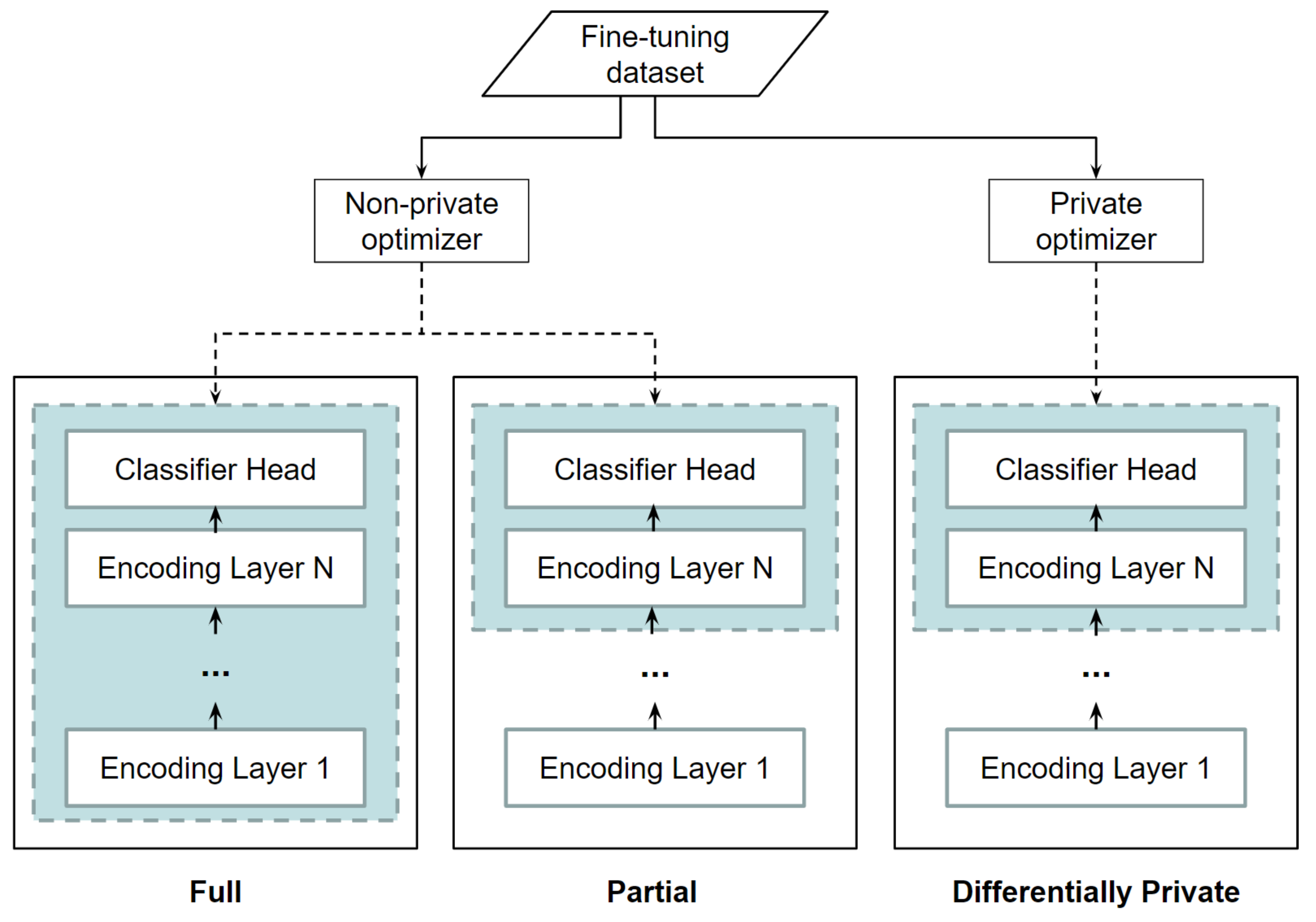}
    \caption[Fine-tuning methods]{
    An illustration of the different fine-tuning methods.}
    \label{fig:fine-tuning}
\end{figure}

In the fine-tuning phase, we employ single-label text-classification as the downstream task. Our setup consists of three different fine-tuning methods: \textit{Full}, \textit{Partial}, and \textit{Differentially Private (DP)} fine-tuning.
The different fine-tuning methods are depicted in Figure~\ref{fig:fine-tuning}.

The \textit{Full} setup follows the standard practices of fine-tuning LMs, where a classifier head is attached to the base network and all the weights of a pre-trained network along with the classifier head are retrained on the task-specific dataset with a low learning rate~\cite{howard2018universal}. Full fine-tuning usually leads to the best results on the downstream task, but in the case of large LMs, it is not always feasible due to the size of these networks and the computational costs of retraining them. Due to this constraint, researchers have designed alternative fine-tuning strategies where fine-tuning is employed in a more optimized manner~\cite{howard2018universal,lee2019would}.
A common alternative strategy is to freeze most of the layers in a network and only retrain the last few encoder layers with the task-specific head of the network \cite{liu2020exploring,sun2022unfreeze}. 
In our \textit{Partial} setup, we freeze all layers of the BERT model except for the last encoding layer. Applying DP in fine-tuning puts additional noise to the gradient updates, which in the lower layers carries a detrimental effect to the pre-trained knowledge of the model as the weights of the bottom layers are more sensitive to noise. For this reason, in the \textit{Differentially Private} setup we employ the same layer-freezing approach as in the \textit{Partial} setup.

\label{sec:dpsgd}
In addition, the DP fine-tuning method uses the Adam variant of DPSGD by Abadi et al.~\cite{abadi2016deep}. DPSGD is a modification to the stochastic gradient descent algorithm that employs ($\mathbf{\epsilon , \delta }$) differential privacy~\cite{dwork2006our}. In the formal definition of ($\mathbf{\epsilon , \delta }$) differential privacy, a randomized algorithm M is differentially private if:
\[
\text{Pr[M(x)}\in\text{S]}\leq\text{exp}(\mathbf{\epsilon}) \cdot \text{Pr[M(y)}\in\text{S]}+\mathbf{\delta}
\]
where x and y are neighboring datasets, S denotes all the potential output of M that can be predicted, $\mathbf{\epsilon}$ is the metric of privacy loss (also known as the privacy budget) at a differential change in the data (\eg adding or removing one datapoint), and $\mathbf{\delta}$ is the probability of an accidental privacy leak. 
In deep learning, an $\mathbf{\epsilon}$ value is defined as modest when it is below 10 and $\mathbf{\delta}$ is usually set to the reciprocal of the number of training samples~\cite{dupuy2021efficient,yu2021differentially}.
In standard data analytics settings, $\mathbf{\epsilon}$ values between 0 and 1 are considered to be highly private, and values between 2 and 10 are considered somewhat private. 
However, in deep learning it is hard to achieve a $\mathbf{\epsilon}$ value under 1, since the
privacy budget is based on how much the data affected the model. Everytime the dataset goes through the model, the $\mathbf{\epsilon}$ increases.

The DPSGD algorithm entails two major changes to the gradient descent algorithm: the introduction of gradient clipping and the addition of Gaussian noise to the clipped gradients. 
The clipping limits how much an individual training point can impact the model parameters, while the addition of the noise randomizes the behavior of the algorithm, making it statistically impossible to know whether or not a training point was included in the training set. 
These modifications to the gradients happen on a microbatch level and are aggregated for the standard batch optimization step.

\subsection{Text Generation}
Although the standard objective of NLG is to produce text that appears indistinguishable from human-written text (see related work in Section~\ref{sec:nlg}), in our study we are less interested in general text quality in terms of coherence or grammatical correctness. 
Our primary goal is to trigger the fine-tuned models to generate named entities found in the training data.
To achieve this, we employ two different prompting methods and an efficient generation strategy that produces diverse text samples.

\subsubsection{Prompt Selection}
Selecting good prompts is a crucial step in triggering the model to unveil information about the training data. 
We employ two different prompting strategies for text generation. 

\begin{enumerate}
\item In the first one, we take the strategy shown to achieve the best results in the experiments of Carlini et al.~\cite{carlini2021extracting}, in which a fixed length substring is randomly sampled as prompt from the Crommon Crawl\footnote{http://commoncrawl.org} dataset. This we refer to as a \textit{naive prompting}, since we randomly use text samples scraped from the internet. 
The selected prompts likely have no or only very little connection with the text and named entities from the fine-tuning dataset.

\item In the second strategy, we create the prompts by randomly selecting sequences of a fixed length from the test set of the fine-tuning data. This setup is considered as \textit{informed prompting} given that the prompts come from the same domain and are generally highly similar to the training data.
\end{enumerate}

\subsubsection{Text Generation}
Despite the fact that the bidirectional nature of BERT does not naturally admit to sequential sampling, Wang and Cho~\cite{wang2019bert} have shown that it is also possible to utilize this strategy for BERT. Although their results suggest that their non-sequential iterative method produces slightly more coherent text than sequential sampling, it requires multiple iterations for each token. Since text coherence is not our primary goal, we choose to employ a computationally less expensive sequential sampling method. 
In this method, we choose a randomly selected prompt (see above) as a seed sequence and extend it with a masked token. 
For each iteration, we predict the masked token and replace the mask.
We add an additional masked token to the extended sequence until we reach the defined sequence length.

On top of this generation method, we also employ a combination of beam search and nucleus sampling as an additional decoding strategy. 
Beam search is a commonly used decoding method in machine translation tasks~\cite{freitag2017beam}. 
Compared to greedy search, where at each iteration only the token with the highest probability is selected, beam search selects multiple tokens at each iteration.
The number of tokens is defined by the beam width parameter and an additional conditional probability is used to construct the best combination of these tokens in a sequence. 
While both greedy search and beam search select tokens based on maximum likelihood, sampling from the probability distribution is also a viable approach. 
The most popular sampling method is top-$k$ sampling, where in each iteration a token is sampled from a set of $k$ candidates with the highest probability. 
Nucleus sampling (also called top-$p$ sampling) is an alternative strategy to top-$k$, where instead of using a set with a fixed length, the smallest possible set is constructed with tokens whose cumulative probability exceeds the probability value of parameter $p$~\cite{holtzman2019curious}. These sampling methods can be combined with both search algorithms.

\subsection{Evaluating Named Entity Memorization}
\label{sec:methods_eval}
Named entities refer to objects and instances that identify one item from a set of other items sharing similar attributes. They usually include entity types like (person) names, organizations, locations, products, and special temporal or numerical expressions like dates or amounts of money~\cite{nadeau2007survey}. 
In order to evaluate the extent to which the models have memorized the named entities found in the fine-tuning dataset, we first extract them from the datasets using Named Entity Recognition (NER) and create a dictionary with the entities and their corresponding entity types. 

After this, we create three different entity sets from this dictionary. 
The first one (\textit{All}) consists of every entity with a character length greater than 3. In the second (\textit{Private}), we do a cross-check of the first entity set with the pre-training data, and remove all entities also present in the pre-training datasets, leaving us a set of entities that only appear in the fine-tuning data. 
For the third and final set (\textit{Private $1$-eidetic}), we keep all $1$-eidetic entities, \ie entities which appeared only once in the fine-tuning dataset, of the second set and discard everything else. 
Once we have these three sets and the generated samples from each model set up (including the samples generated from a base model that was not fine-tuned), we count the number of exact matches in the samples.

\section{Experimental Apparatus}
\label{sec:experiments}

\subsection{Datasets}
\label{sec:datasets}
For selecting suitable datasets for the study we had two criteria. First, to avoid privacy issues and ethical concerns only publicly available datasets were chosen. Second, to provide a good basis for the measurement of memorization, we were interested in datasets that contain a large number of named entities. We choose data which has English as its primary language and kept 20\% of each dataset for testing. %
The main characteristics of the datasets can be seen in Table~\ref{tab:datasets}.

\begin{table}[ht]
   \small
    \centering
    \caption{Characteristics of the datasets}\label{tab:datasets}
    \begin{tabular}{lrrrr}
    \toprule
    Dataset & N       & \#Train & \#Test  & \#Classes \\  
    \midrule                                                           
    Enron           & 7,501  & 6,000  & 1,501  & 7        \\ 
    BlogAuthorship  & 430,269 & 344,215 & 86,054  & 39    \\   
    \bottomrule
    \end{tabular}
\end{table}

\subsubsection{Enron Email Dataset}
The raw Enron Email corpus~\cite{klimt2004introducing} consists of 619,446 email messages from 158 employees of the Enron corporation. The dataset contains full emails of real users, which include naturally occurring personal information such as names, addresses, organizations, social security numbers etc.
Since the original dataset is not fitted to text-classification (as it lacks any official labels), we adapted it by labeling the emails by the folder names they are attached to (\ie ``sent-mail'', ``corporate'' , ``junk'', ``proposals'' etc.).
We selected seven folder that could be considered as valid classes in an applied setting. The labels of these seven classes are ``logistics'', ``personal'', ``management'', ``deal discrepancies'', ``resumes'', ``online trading'', and ``corporate''.
During the preprocessing of the emails, we removed the forward blocks, HTML links, line breaks, and tabs. The removal of forward blocks and HTML links was especially important to improve the quality of the generated texts.

\subsubsection{Blog Authorship Corpus}

The Blog Authorship Corpus~\cite{schler2006effects} contains text from blogs written until 2004, with each blog being the work of a single user. The corpus incorporates a total of 681,288 posts from 19,320 users. 
Alongside the blogposts, the dataset includes topic labels and demographic information about the writer, including gender, age, and zodiac sign. 
Although the blogposts were written for the public, they contain some PII such as names, organizations and postal addresses. We adapt the dataset for text classification by labeling the posts by the blogs' topics.
Posts with the topic label ``unknown'' were removed. 
After the removal, the dataset consists of 430,269 posts with 39 unique labels. Preprocessing was kept to a minimal: 
non-printable ASCII characters,
non-ASCII characters (\eg Korean letters), 
and URL links were removed, otherwise the text remained unchanged.

\subsection{Procedure \& Implementation}

The procedure of our experiments follows the pipeline as illustrated in Figure~\ref{fig:pipeline}.
Below, we describe the details of each step.
All experiments were conducted on a NVIDIA A100 HGX GPU with 40~GB of RAM.

\subsubsection{Fine-tuning}
The experiments are based on the Hugging Face implementation of the BERT base uncased model~\cite{wolf2020transformers}. For single-label classification, a custom classifier head is attached to the base model consisting of a Dropout and a Linear layer. In the \textit{Full} and \textit{Partial} setup we used the standard Adam optimizer, while in the \textit{Differentially Private} fine-tuning, we changed it to the DPAdam optimizer from the Opacus library~\cite{yousefpour2021opacus} with a microbatch size of~1.

\subsubsection{Text Generation}
For text generation, we first removed the classifier heads from the fine-tuned models and attached a pre-trained MLM head instead. We then used the sequential generation method described in Section~\ref{sec:methods}, with the addition of beam search and nucleus sampling combined with a temperature parameter. During prompt creation, we sampled a string with a character length of 100 either from the Common Crawl dataset (naive prompting) or from the test set (informed prompting). 
We set the sequence length to 256 tokens.
We removed the tokens of the prompt before saving the samples, \ie if the prompt contained any entities, they are not considered for the evaluation. 
In total, we generated 20,000 text samples for each setup.

\subsubsection{Named Entity Recognition}
\label{sec:procedure_ner}
For collecting the named entities from the fine-tuning datasets, we employed the NER system of the spaCy library that utilizes a custom word embedding strategy, a transformer, and a transition-based approach to named entity parsing~\cite{zenodo121230}. 
spaCy distinguishes between a total of 18 different named entity types. 
Out of these 18 entity types we selected seven (Person, Organization, Location, Geo-Political Event, Facility, Money, Cardinal), which have a high possibility to contain personal or privacy sensitive information.
When creating the \textit{Private} entity set described in Section~\ref{sec:methods_eval}, we cross-checked our fine-tuning entities  with the pre-training datasets (the Book Corpus and Wikipedia datasets, available through the datasets library~\cite{lhoest-etal-2021-datasets}) to discard the entities present both in fine-tuning and pre-training. 
The numbers of named entities per type in each of the three sets can be seen in Table~\ref{tab:named_ents}.

\setlength{\tabcolsep}{1pt}
\begin{table*}[!th]
    \small
    \centering
    \caption[Named entity recognition results]{Number of named entities found in the datasets sorted by type. 
    }\label{tab:named_ents}
    \begin{tabular}{lr @{\extracolsep{0.3cm}} rr|rrr}
    \toprule
    Named Entity Type & \multicolumn{3}{c}{Enron} & \multicolumn{3}{c}{Blog Authorship}\\  
    \midrule
    & All & Private & Private     & All & Private & Private\\
    &     &         & $1$-eidetic &     &         & $1$-eidetic   \\
    \hline
    PERSON       & 10,712 & 7,717 & 4,844 & 209,434 & 137,892 & 113,599 \\ 
    ORG          & 9,933 & 7,178 & 5,001 & 168,068 & 107,480 & 90,594 \\
    LOC          & 316 & 175 & 125 & 10,562 & 4,902 & 4,049 \\
    GPE          & 1,551 & 739 & 490 & 37,691 & 17,781 & 13,196 \\
    FAC          & 367 & 230 & 174 & 12,824 & 7,137 & 6,349 \\
    MONEY        & 1,220 & 736 & 585 & 11,216 & 7,551 & 6,343 \\
    CARDINAL     & 2,918 & 1,924 & 1,386 & 24,075 & 13,020 & 10,810 \\
    \hline
    Total & 27,017 & 18,726 & 12,605 & 473,870 & 295,763 & 244,940 \\
    \bottomrule
    \end{tabular}
\end{table*}
\subsection{Hyperparameter Optimization}

\subsubsection{Fine-tuning}

During fine-tuning, we carefully optimized the models on both datasets using manual tuning based on test accuracy. 
Dropout rates were fixed based on the default BERT base implementation of the huggingface library (0.1 for attention dropout and 0.3 for the classifier)~\cite{wolf2020transformers}. For batch size, learning rate, and number of epochs a search space was defined based on previous works~\cite{DBLP:conf/acl/GalkeS22,dodge2020fine}. Specifically, we chose the batch size from \{8, 16, 32\}, the learning rate from \{5e-3, 1e-3, 1e-4, 5e-5, 1e-5\}, and the number of epoch from \{3, 5, 10\}.
Across all setups we found that using a batch size of 32 leads to the best performance. On the Enron dataset the highest accuracy values were achieved when the number of epochs is set to 10, while the Blog Authorship dataset required 5 epochs to reach the highest values.
In the \textit{Full} setup a learning of 1e-5 was found to be the best performing on both datasets. In the \textit{Partial} setup the best results were found with a 5e-5 learning rate for the Enron dataset and 1e-4 for the Blog Authorship dataset. 




In the \textit{DP} fine-tuning, the best results were achieved with a learning rate of 1e-3 on both datasets. In this setup two additional hyperparameters had to be optimized to achieve the highest possible accuracy while keeping the privacy budget $\epsilon$ single-digit. We set the per-example gradient clipping threshold to 10 based on a previous study using DP with BERT~\cite{yu2021differentially}, and found the best values for the noise multiplier to be 0.5 for the Enron and 0.4 for the Blog Authorship dataset.
\subsubsection{Text Generation}
For text generation, we studied the effects of the different sampling parameters.
We found the best results in terms of text diversity and coherence through manual tuning with the following value combinations:
number of beams: 1,
beam size: 30,
nucleus sampling value: 0.8,
temperature: 2.0, and
n-gram repetition limit: 3.

\subsection{Measures}

To evaluate the performance on the downstream task, we use accuracy.
%
In the \textit{DP} setup, we measure privacy preservation with the privacy budget $\mathbf{\epsilon}$. 
In all models, the extent of unintended memorization of named entities found in the fine-tuning dataset is measured by counting their occurrences in generated samples and checking their $k$-eidetic value.
A data point (or in our case an entity) is $k$-eidetic if it appears $k$ times in the training corpus~\cite{carlini2021extracting}.

\section{Results}
\label{sec:results}
\subsection{Classification}
Table~\ref{tab:accuracy} shows the singe-label text classification results for each setup. 
For both datasets, we observe a similar trend between the different fine-tuning setups: \textit{Full} achieved the highest accuracy on the test set, \textit{Partial} performed slightly worse, and the \textit{DP} setup produced the worst results with a $16$ percent point drop compared to the \textit{Partial}. 
In general, the accuracy values are considerably higher on the Enron dataset. In the \textit{DP} setup, the privacy budget $\mathbf{\epsilon}$ is $9.79$ for the Enron and $\epsilon=7.38$ for Blog Authorship.

\begin{table}[h]
   \small
    \centering
    \caption{Mean accuracy and standard deviation over five runs on the single-label text classification}
    \begin{tabular}{l @{\extracolsep{0.5cm}} r @{\extracolsep{0.5cm}} r}
    \toprule
    Fine-tuning Setup & Enron & Blog Authorship \\  
    \midrule                                                    
    Full                            & 86.83\% (0.46) & 51.69\% (0.32)\\ 
    Partial                         & 85.95\% (0.34)& 49.58\% (0.15)\\
    DP                              & 68.28\% (0.88) & 35.86\% (0.12)\\
    \bottomrule
    \end{tabular}
    \label{tab:accuracy}
\end{table}

\begin{figure*}[h]
    \centering
    \includegraphics[width=12cm, height=4.5cm]{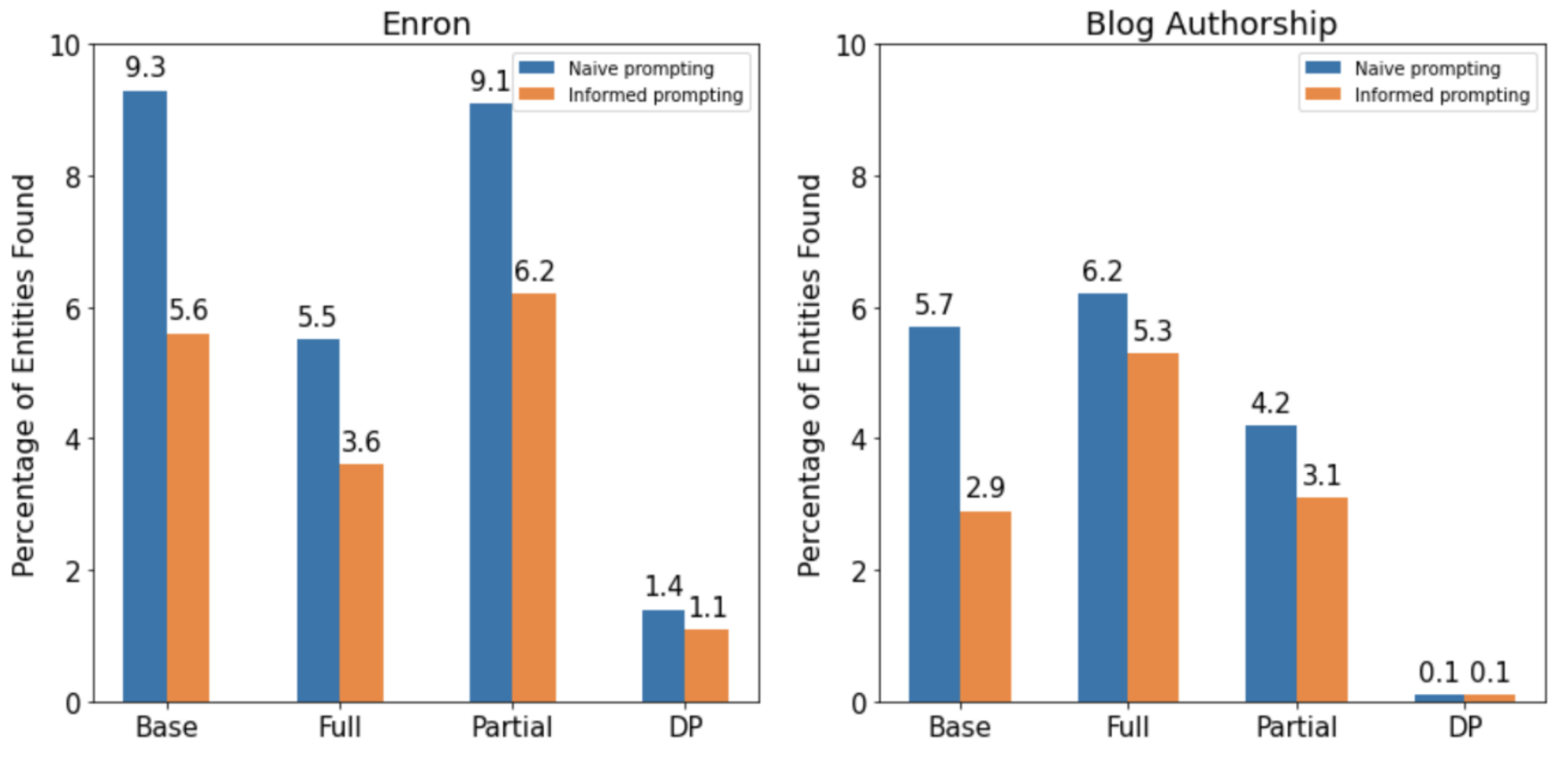}
    \caption[Memorization rates for all entities]{The percentages of all entities successfully extracted from the models, compared by prompting methods.}
    \label{fig:all_ent}
\end{figure*}

\begin{figure*}[ht]
    \centering
    \includegraphics[width=12cm, height=5cm]{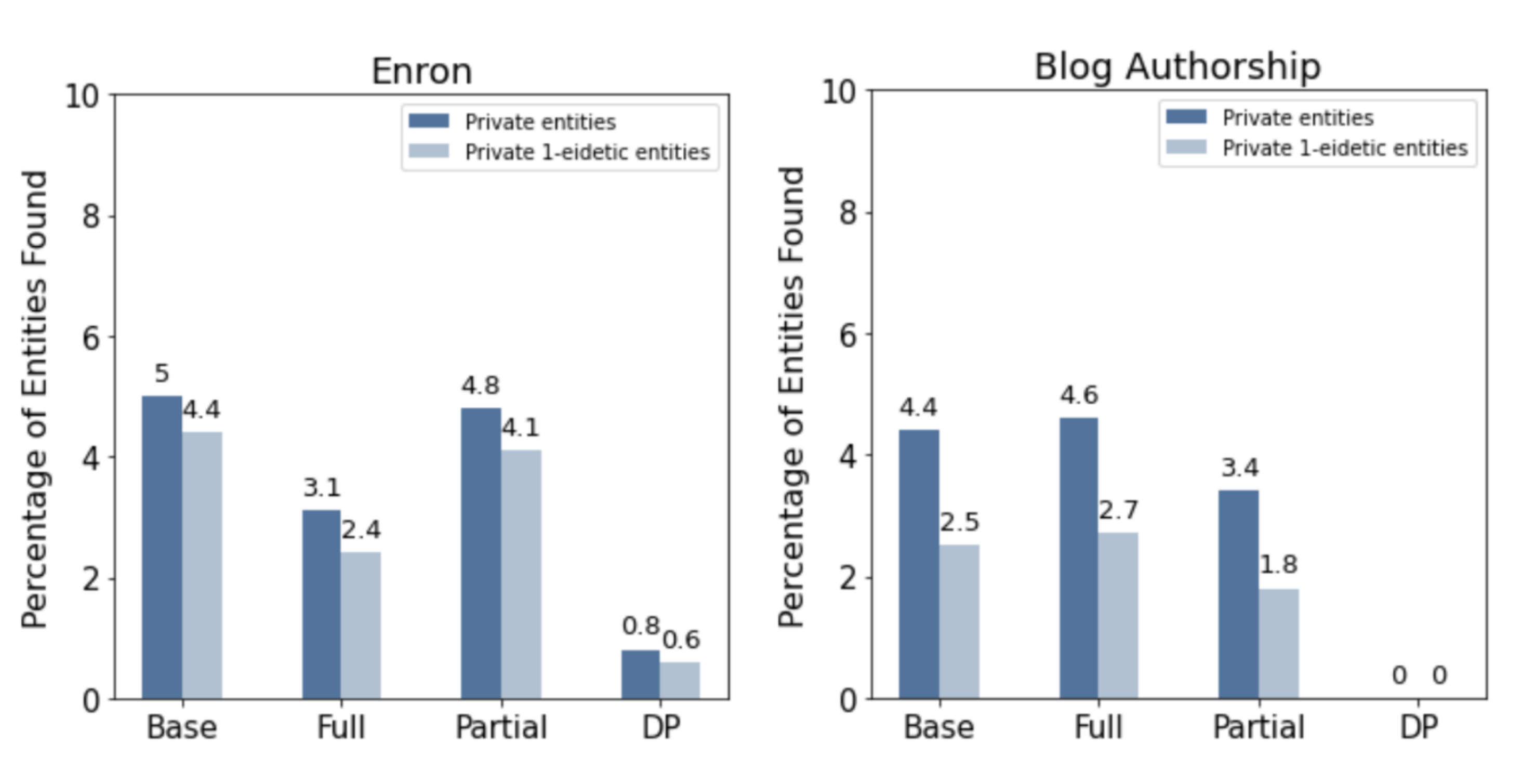}
    \caption[Memorization rates for private named entities and private $1$-eidetic named entities]{The percentages of private entities and private $1$-eidetic entities successfully extracted from the models with the use of naive prompting.}
    \label{fig:combined_ent}
\end{figure*}

\begin{table*}[ht]
    \small
    \caption[Memorization rates for private named entity types]{Extraction ratio of entities from the \textit{Private} set using naive prompting, grouped by entity types}
    \centering
    \begin{tabular}{lr @{\extracolsep{0.25cm}} rrr|rrrr}
    \toprule
    Named Entity Type & \multicolumn{4}{c}{Enron} & \multicolumn{4}{c}{Blog Authorship}\\  
    \midrule
      & Base & Full & Partial & DP & Base & Full & Partial & DP   \\
    \hline
    PERSON       & 4.1\% & 2.3\% & 4.3\% & 0.8\% & 3.4\% & 4.1\% & 2.9\% & *\\ 
    ORG          & 3.8\% & 2.2\% & 3.3\% & 0.3\% & 4.5\% & 4.1\% & 3\%& *\\
    LOC          & 20.5\% & 15.4\% & 18.4\% & 5.1\% & 8.9\% & 8.6\% & 5.5\% & *\\
    GPE          & 28.1\% & 22.5\% & 28\% & 5\% & 11.5\% & 13.4\% & 9.9\% & 0.1\% \\
    FAC          & 1.7\% & 0.4\% & 0.8\% & 0.8\% & 2.7\% & 2.5\% & 1.6\% & *\\
    MONEY        & 1.5\% & 0.7\% & 1\% & 0.1\% & 1.2\% & 0.9\% & 0.7\% & *\\
    CARDINAL     & 4.8\% & 1.7\% & 3.9\% & 0.5\% & 4.4\% & 3.9\% & 2.7\% & 0.1\% \\
    \bottomrule
    * less than 0.1\%
    \end{tabular}
    \label{tab:extracted_ents}
\end{table*}
\subsection{Named Entity Memorization}
For the named entity memorization experiments, we also included a pre-trained only BERT, \ie without any fine-tuning, which we call the \textit{Base} setup.
Figure~\ref{fig:all_ent} shows our initial results on the \textit{All} entity set. 
The highest extraction rate was 9.3\% for the Enron and 6.2\% for the Blog Authorship dataset. 
On the Enron dataset, the highest extraction rate was achieved on the \textit{Base} setup. closely followed by the \textit{Partial} setup. The difference between these two setups was $0.2\%$ with the naive prompting and $0.6\%$ for the informed prompting methods.
On the Blog Authorship dataset, the \textit{Full} setup produced the highest extraction rate, followed by the \textit{Base} setup. 
Between these two setups, the naive prompting resulted in $0.5\%$  and the informed prompting in a $2.4\%$ difference. 
The \textit{DP} setup produced the lowest extraction rates with $1.4\%$ (naive prompting) and $1.1\%$ (informed prompting) on the Enron and $0.1\%$ in the Blog Authorship dataset (both, naive and informed prompting). 
Naive prompting consistently outperformed informed prompting in all fine-tuning setups on both datasets.

Figure~\ref{fig:combined_ent} shows the comparison of the  extraction rates between the private entities and the private $1$-eidetic entities, using the naive prompting method. 
Compared to the results in Figure~\ref{fig:all_ent}, the extraction rates are consistently lower across all setups. The difference between \textit{Base} and \textit{Partial} on Enron, and \textit{Base} and \textit{Full} on Blog Authorship once again is negligible. 
Overall the memorization rate of private $1$-eidetic entities is lower than the memorization rate of all private entities.
But the difference is less than $1$ percent point on the Enron and less than $2$ percent points on the Blog Authorship dataset.

To further investigate the extracted private entities, we also measured the extraction ratio of each entity type in Table~\ref{tab:extracted_ents}. 
The Location and Geo-Political Event types produced the highest percentages, while the Facility and Money types had results less than $3\%$ across all setups and datasets. The extraction ratios on Blog Authorship are consistently lower in every entity type compared to Enron. The only exception can be seen in the Facility type, where the Blog Authorship results were 1 to 2 percent points higher.

\section{Discussion}
\label{sec:discussion}
\subsection{Key Insights}
\label{sec:key}

\paragraph{\textbf{Prompting Methods}}
Our experiments show that the naive prompting method produces better results in all setups. 
Although for informed prompting the seed sequences will be more similar to the text sequences found in the fine-tuning data, this informed prompting likewise limits the possibilities of producing diverse outputs. 
Following Carlini et al.~\cite{carlini2021extracting}, we conclude that using random prompts sampled from a huge corpus unrelated to the training data yields better extraction results. 
This shows that adversaries do not need to have prior knowledge about the training data of the attacked model, a simple black-box approach is sufficient.

\paragraph{\textbf{Named Entity Memorization in BERT}}
We extracted private named entities from the fine-tuned models at surprisingly low rates.
In no setup, we extracted more than $10\%$ of the private entities.
Interestingly, our results further show that using a pre-trained \textit{Base} model that has not been fine-tuned on the training set containing those extracted entities produces similar extraction ratios. 
Our assumption is that the small percentage of private entities that have been successfully extracted from both the \textit{Base} and \textit{Full} or \textit{Partial} models have low level of complexity in terms of length and n-gram diversity. 
Therefore, they are more likely to be randomly generated by combining common subword tokens.

In order to better understand the reasons behind this observation, we conducted a more detailed analysis of the extracted entities. 
As can be seen from Table~\ref{tab:extracted_ents}, distinct entity types have different probabilities to be extracted. 
From the seven types, we argue Location and Geo-Political Event are the least unique in their nature, therefore it is not suprising that the highest extraction rates have been achieved on them. The lower values in the Money and Cardinal types reinforce the findings that the subword tokenization in BERT is a suboptimal method to encode numerical values~\cite{wallace2019nlp}. 
Overall our findings suggest that BERT could be rather resistant to training data extraction attacks unlike other large LMs such as GPT-2 \cite{carlini2021extracting}. 
This is most likely due to its smaller size as argued in~\cite{carlini2021extracting}.
It is also possible that auto-encoder transformers are generally less prone to these attacks compared to auto-regressive transformers as result of their different pre-training objectives.

\paragraph{\textbf{Differentially Private Fine-tuning}}
In all the experiments that used Differentially Private fine-tuning, the extraction rates of named entities were reduced by a large extent. 
Our samples have shown that the text quality in the \textit{DP} setups was very low, both text coherence and text diversity decreased dramatically. 
Even though the performance on the downstream task was also considerably lower, we argue this trade-off between performance and privacy is still promising for future developments. 
Considering that the focus of our study was not on achieving state-of-the art performance for singe-label text classification, we only used the Adam variant of the original DPSGD algorithm~\cite{abadi2016deep}. 
We leave the use of other, more advanced DP algorithms 
like~\cite{yu2021differentially} to future work. 
One can expect that for tasks, where the ability of a model to generate text is irrelevant, the use of DP can be a viable solution to increase the privacy of the model.

\subsection{Generalization}
\label{sec:generalization}
In our experiments, we intentionally used datasets of different characteristics.
While the Enron dataset we used is small in size and is very cluttered due to its source (real world emails), the Blog Authorship Corpus is a public web corpus that contains a large amount of samples covering a broad range of domains with a higher text quality. 
Although, we only used single-label text classification (in which BERT is generally considered as state-of-the-art~\cite{DBLP:conf/acl/GalkeS22}) as a downstream task for fine-tuning, results should be similar on different downstream tasks since the memorization takes place in the encoding layers, irrespectively of the task-specific final layers of a model.
Finally our conclusion about the memorization capabilities of the BERT base model is in line with the training data extraction study done on Clinical BERT, in which the authors were unable to reliably extract patient names from a specific BERT variant pre-trained on clinical data~\cite{lehman2021does}.

\subsection{Threats to Validity}
\label{sec:threattovalidity}
We acknowledge that the experimental datasets are limited to English. Although named entities are often unique to their respective language, we have no reason to believe that generating named entities would be significantly easier in other languages. 
For languages that have larger character sets (\eg Chinese) or use long compound words (\eg German), the probability of unintended memorization may even be smaller. Regarding the efficiency of the extraction of named entities, the results can be influenced by both the named entity recognition system and our text generation method. It is  possible that some entities have been missed and some have been falsely identified. The missed entities are unlikely to influence the results since we still had a great amount of entities of differing $k$-eidetic values. Controlling for the falsely identified entities was a more difficult problem.
Therefore, we decided to remove all entities with a character length of less than 4. 
Using a left-to-right sequential text generation method 
might also bias our results, as BERT uses context from both directions to predict a token during pre-training. 
This, we argue has more impact on text coherence rather than the ability to trigger a diverse output containing named entities.
The latter was of higher importance to our study.

\section{Conclusion}
\label{sec:conclusion}
As the capabilities of large LMs increase, it is important that the privacy aspects of these models are also considered. We performed an investigation into the capabilities of BERT to memorize named entities. We ran experiments, in which we tried to extract private named entities from fine-tuned BERT models using three different fine-tuning methods and two prompting strategies.
Overall, we could only extract a low percentage of named entities from BERT, and found that the pre-trained only model generates the same amount of entities as the fine-tuned models. We also employed a Differential Private fine-tuning method, which showed to be a promising privacy preserving method against training data extraction attacks with some trade-off on the downstream task performance. Although our results do not rule out the possibility to extract personal information from a fine-tuned BERT base model using more advanced methods, our findings suggest that doing so is at least not trivial. As for future work it would be interesting to re-run the experiments on other commonly used language models and to test the embedding layers of our BERT setups against membership inference attacks.
%
%

%
%
%
\newpage
\bibliographystyle{splncs04}
\bibliography{library}

\begin{thebibliography}{10}
\providecommand{\url}[1]{\texttt{#1}}
\providecommand{\urlprefix}{URL }
\providecommand{\doi}[1]{https://doi.org/#1}

\bibitem{abadi2016deep}
Abadi, M., Chu, A., Goodfellow, I., McMahan, H.B., Mironov, I., Talwar, K.,
  Zhang, L.: Deep learning with differential privacy. In: Proceedings of the
  2016 ACM SIGSAC conference on computer and communications security. pp.
  308--318 (2016)

\bibitem{acar2018survey}
Acar, A., Aksu, H., Uluagac, A.S., Conti, M.: A survey on homomorphic
  encryption schemes: Theory and implementation. ACM Computing Surveys (CSUR)
  \textbf{51}(4),  1--35 (2018)

\bibitem{alsentzer2019publicly}
Alsentzer, E., Murphy, J.R., Boag, W., Weng, W.H., Jin, D., Naumann, T.,
  McDermott, M.: Publicly available clinical bert embeddings. arXiv preprint
  arXiv:1904.03323  (2019)

\bibitem{aono2017privacy}
Aono, Y., Hayashi, T., Wang, L., Moriai, S., et~al.: Privacy-preserving deep
  learning via additively homomorphic encryption. IEEE Transactions on
  Information Forensics and Security  \textbf{13}(5),  1333--1345 (2017)

\bibitem{beltagy2019scibert}
Beltagy, I., Lo, K., Cohan, A.: Scibert: A pretrained language model for
  scientific text. arXiv preprint arXiv:1903.10676  (2019)

\bibitem{brown2020language}
Brown, T., Mann, B., Ryder, N., Subbiah, M., Kaplan, J.D., Dhariwal, P.,
  Neelakantan, A., Shyam, P., Sastry, G., Askell, A., et~al.: Language models
  are few-shot learners. Advances in neural information processing systems
  \textbf{33},  1877--1901 (2020)

\bibitem{carlini2019secret}
Carlini, N., Liu, C., Erlingsson, {\'U}., Kos, J., Song, D.: The secret sharer:
  Evaluating and testing unintended memorization in neural networks. In: 28th
  USENIX Security Symposium (USENIX Security 19). pp. 267--284 (2019)

\bibitem{carlini2021extracting}
Carlini, N., Tramer, F., Wallace, E., Jagielski, M., Herbert-Voss, A., Lee, K.,
  Roberts, A., Brown, T., Song, D., Erlingsson, U., et~al.: Extracting training
  data from large language models. In: 30th USENIX Security Symposium (USENIX
  Security 21). pp. 2633--2650 (2021)

\bibitem{davody2020robust}
Davody, A., Adelani, D.I., Kleinbauer, T., Klakow, D.: Robust differentially
  private training of deep neural networks. arXiv preprint arXiv:2006.10919
  (2020)

\bibitem{de2020overview}
De~Cristofaro, E.: An overview of privacy in machine learning. arXiv preprint
  arXiv:2005.08679  (2020)

\bibitem{devlin2018bert}
Devlin, J., Chang, M.W., Lee, K., Toutanova, K.: {BERT}: Pre-training of deep
  bidirectional transformers for language understanding. arXiv preprint
  arXiv:1810.04805  (2018)

\bibitem{dodge2020fine}
Dodge, J., Ilharco, G., Schwartz, R., Farhadi, A., Hajishirzi, H., Smith, N.:
  Fine-tuning pretrained language models: Weight initializations, data orders,
  and early stopping. arXiv preprint arXiv:2002.06305  (2020)

\bibitem{dupuy2021efficient}
Dupuy, C., Arava, R., Gupta, R., Rumshisky, A.: An efficient {DP-SGD} mechanism
  for large scale {NLP} models. arXiv preprint arXiv:2107.14586  (2021)

\bibitem{dwork2006our}
Dwork, C., Kenthapadi, K., McSherry, F., Mironov, I., Naor, M.: Our data,
  ourselves: Privacy via distributed noise generation. In: Annual international
  conference on the theory and applications of cryptographic techniques. pp.
  486--503. Springer (2006)

\bibitem{dwork2014algorithmic}
Dwork, C., Roth, A., et~al.: The algorithmic foundations of differential
  privacy. Found. Trends Theor. Comput. Sci.  \textbf{9}(3-4),  211--407 (2014)

\bibitem{fredrikson2015model}
Fredrikson, M., Jha, S., Ristenpart, T.: Model inversion attacks that exploit
  confidence information and basic countermeasures. In: Proceedings of the 22nd
  ACM SIGSAC conference on computer and communications security. pp. 1322--1333
  (2015)

\bibitem{freitag2017beam}
Freitag, M., Al-Onaizan, Y.: Beam search strategies for neural machine
  translation. In: Proceedings of the First Workshop on Neural Machine
  Translation. pp. 56--60 (2017)

\bibitem{DBLP:conf/acl/GalkeS22}
Galke, L., Scherp, A.: Bag-of-words vs. graph vs. sequence in text
  classification: Questioning the necessity of text-graphs and the surprising
  strength of a wide {MLP}. In: Proceedings of the 60th Annual Meeting of the
  Association for Computational Linguistics (Volume 1: Long Papers), {ACL}
  2022, May 22-27, 2022. pp. 4038--4051. Association for Computational
  Linguistics (2022)

\bibitem{geman1984stochastic}
Geman, S., Geman, D.: Stochastic relaxation, gibbs distributions, and the
  bayesian restoration of images. IEEE Transactions on pattern analysis and
  machine intelligence (6),  721--741 (1984)

\bibitem{ghazvininejad2019mask}
Ghazvininejad, M., Levy, O., Liu, Y., Zettlemoyer, L.: Mask-predict: Parallel
  decoding of conditional masked language models. arXiv preprint
  arXiv:1904.09324  (2019)

\bibitem{gilad2016cryptonets}
Gilad-Bachrach, R., Dowlin, N., Laine, K., Lauter, K., Naehrig, M., Wernsing,
  J.: Cryptonets: Applying neural networks to encrypted data with high
  throughput and accuracy. In: International conference on machine learning.
  pp. 201--210. PMLR (2016)

\bibitem{hassan2018anonymization}
Hassan, F., Domingo-Ferrer, J., Soria-Comas, J.: Anonymization of unstructured
  data via named-entity recognition. In: International conference on modeling
  decisions for artificial intelligence. pp. 296--305. Springer (2018)

\bibitem{he2020deberta}
He, P., Liu, X., Gao, J., Chen, W.: De{BERT}a: Decoding-enhanced {BERT} with
  disentangled attention. arXiv preprint arXiv:2006.03654  (2020)

\bibitem{holtzman2019curious}
Holtzman, A., Buys, J., Du, L., Forbes, M., Choi, Y.: The curious case of
  neural text degeneration. arXiv preprint arXiv:1904.09751  (2019)

\bibitem{zenodo121230}
Honnibal, M., Montani, I., Van~Landeghe, S., Boyd, A.: spacy:
  Industrial-strength natural language processing in python (2022),
  \url{https://zenodo.org/record/121230}

\bibitem{howard2018universal}
Howard, J., Ruder, S.: Universal language model fine-tuning for text
  classification. arXiv preprint arXiv:1801.06146  (2018)

\bibitem{klimt2004introducing}
Klimt, B., Yang, Y.: Introducing the enron corpus. In: {CEAS} 2004 - First
  Conference on Email and Anti-Spam, July 30-31, 2004, Mountain View,
  California, {USA} (2004)

\bibitem{lee2019would}
Lee, J., Tang, R., Lin, J.: What would elsa do? freezing layers during
  transformer fine-tuning. arXiv preprint arXiv:1911.03090  (2019)

\bibitem{lehman2021does}
Lehman, E., Jain, S., Pichotta, K., Goldberg, Y., Wallace, B.C.: Does bert
  pretrained on clinical notes reveal sensitive data? arXiv preprint
  arXiv:2104.07762  (2021)

\bibitem{lhoest-etal-2021-datasets}
Lhoest, Q., Villanova~del Moral, A., Jernite, Y., Thakur, A., von Platen, P.,
  Patil, S., Chaumond, J., Drame, M., Plu, J., Tunstall, L., Davison, J.,
  Sasko, M., Chhablani, G., Malik, B., Brandeis, S., Le~Scao, T., Sanh, V., Xu,
  C., Patry, N., McMillan-Major, A., Schmid, P., Gugger, S., Delangue, C.,
  Matussiere, T., Debut, L., Bekman, S., Cistac, P., Goehringer, T., Mustar,
  V., Lagunas, F., Rush, A., Wolf, T.: Datasets: A community library for
  natural language processing. In: Proceedings of the 2021 Conference on
  Empirical Methods in Natural Language Processing: System Demonstrations. pp.
  175--184. Association for Computational Linguistics, Online and Punta Cana,
  Dominican Republic (2021)

\bibitem{liu2021machine}
Liu, B., Ding, M., Shaham, S., Rahayu, W., Farokhi, F., Lin, Z.: When machine
  learning meets privacy: A survey and outlook. ACM Computing Surveys (CSUR)
  \textbf{54}(2),  1--36 (2021)

\bibitem{liu2019roberta}
Liu, Y., Ott, M., Goyal, N., Du, J., Joshi, M., Chen, D., Levy, O., Lewis, M.,
  Zettlemoyer, L., Stoyanov, V.: Ro{BERT}a: A robustly optimized {BERT}
  pretraining approach. arXiv preprint arXiv:1907.11692  (2019)

\bibitem{liu2020exploring}
Liu, Z., Winata, G.I., Madotto, A., Fung, P.: Exploring fine-tuning techniques
  for pre-trained cross-lingual models via continual learning. arXiv preprint
  arXiv:2004.14218  (2020)

\bibitem{mao2020survey}
Mao, H.H.: A survey on self-supervised pre-training for sequential transfer
  learning in neural networks. arXiv preprint arXiv:2007.00800  (2020)

\bibitem{mcmahan2016federated}
McMahan, H.B., Moore, E., Ramage, D., y~Arcas, B.A.: Federated learning of deep
  networks using model averaging. arXiv preprint arXiv:1602.05629  (2016)

\bibitem{mireshghallah2020privacy}
Mireshghallah, F., Taram, M., Vepakomma, P., Singh, A., Raskar, R.,
  Esmaeilzadeh, H.: Privacy in deep learning: A survey. arXiv preprint
  arXiv:2004.12254  (2020)

\bibitem{nadeau2007survey}
Nadeau, D., Sekine, S.: A survey of named entity recognition and
  classification. Lingvisticae Investigationes  \textbf{30}(1),  3--26 (2007)

\bibitem{oh2019towards}
Oh, S.J., Schiele, B., Fritz, M.: Towards reverse-engineering black-box neural
  networks. In: Explainable AI: Interpreting, Explaining and Visualizing Deep
  Learning, pp. 121--144. Springer (2019)

\bibitem{ouyang2022training}
Ouyang, L., Wu, J., Jiang, X., Almeida, D., Wainwright, C.L., Mishkin, P.,
  Zhang, C., Agarwal, S., Slama, K., Ray, A., et~al.: Training language models
  to follow instructions with human feedback. arXiv preprint arXiv:2203.02155
  (2022)

\bibitem{parisot2021property}
Parisot, M.P., Pejo, B., Spagnuelo, D.: Property inference attacks on
  convolutional neural networks: Influence and implications of target model's
  complexity. arXiv preprint arXiv:2104.13061  (2021)

\bibitem{raffel2019exploring}
Raffel, C., Shazeer, N., Roberts, A., Lee, K., Narang, S., Matena, M., Zhou,
  Y., Li, W., Liu, P.J.: Exploring the limits of transfer learning with a
  unified text-to-text transformer. arXiv preprint arXiv:1910.10683  (2019)

\bibitem{rigaki2020survey}
Rigaki, M., Garcia, S.: A survey of privacy attacks in machine learning. arXiv
  preprint arXiv:2007.07646  (2020)

\bibitem{rogers2020primer}
Rogers, A., Kovaleva, O., Rumshisky, A.: A primer in {BERT}ology: What we know
  about how {BERT} works. Transactions of the Association for Computational
  Linguistics  \textbf{8},  842--866 (2020)

\bibitem{rubinstein2009learning}
Rubinstein, B.I., Bartlett, P.L., Huang, L., Taft, N.: Learning in a large
  function space: Privacy-preserving mechanisms for {SVM} learning. arXiv
  preprint arXiv:0911.5708  (2009)

\bibitem{sanh2019distilbert}
Sanh, V., Debut, L., Chaumond, J., Wolf, T.: Distil{BERT}, a distilled version
  of {BERT}: smaller, faster, cheaper and lighter. arXiv preprint
  arXiv:1910.01108  (2019)

\bibitem{schler2006effects}
Schler, J., Koppel, M., Argamon, S., Pennebaker, J.: Effects of age and gender
  on blogging in proceedings of 2006 aaai spring symposium on computational
  approaches for analyzing weblogs. In: Proceedings of 2006 AAAI Spring
  Symposium on Computational Approaches for Analyzing Weblogs (2006)

\bibitem{sharir2020cost}
Sharir, O., Peleg, B., Shoham, Y.: The cost of training {NLP} models: A concise
  overview. arXiv preprint arXiv:2004.08900  (2020)

\bibitem{shokri2017membership}
Shokri, R., Stronati, M., Song, C., Shmatikov, V.: Membership inference attacks
  against machine learning models. In: 2017 IEEE symposium on security and
  privacy (SP). pp. 3--18. IEEE (2017)

\bibitem{DBLP:conf/doceng/SinghoferGKS21}
Singhofer, F., Garifullina, A., Kern, M., Scherp, A.: A novel approach on the
  joint de-identification of textual and relational data with a modified
  mondrian algorithm. In: DocEng '21: {ACM} Symposium on Document Engineering
  2021, August 24-27, 2021. pp. 14:1--14:10. {ACM} (2021)

\bibitem{sun2022unfreeze}
Sun, W., Khan, H., Guenon~des Mesnards, N., Rubino, M., Arkoudas, K.: Unfreeze
  with care: Space-efficient fine-tuning of semantic parsing models. In:
  Proceedings of the ACM Web Conference 2022. pp. 999--1007 (2022)

\bibitem{sweeney2002k}
Sweeney, L.: k-anonymity: A model for protecting privacy. International Journal
  of Uncertainty, Fuzziness and Knowledge-Based Systems  \textbf{10}(05),
  557--570 (2002)

\bibitem{thakkar2020understanding}
Thakkar, O., Ramaswamy, S., Mathews, R., Beaufays, F.: Understanding unintended
  memorization in federated learning. arXiv preprint arXiv:2006.07490  (2020)

\bibitem{tramer2016stealing}
Tram{\`e}r, F., Zhang, F., Juels, A., Reiter, M.K., Ristenpart, T.: Stealing
  machine learning models via prediction apis. In: 25th USENIX security
  symposium (USENIX Security 16). pp. 601--618 (2016)

\bibitem{vaswani2017attention}
Vaswani, A., Shazeer, N., Parmar, N., Uszkoreit, J., Jones, L., Gomez, A.N.,
  Kaiser, {\L}., Polosukhin, I.: Attention is all you need. Advances in neural
  information processing systems  \textbf{30} (2017)

\bibitem{wallace2019nlp}
Wallace, E., Wang, Y., Li, S., Singh, S., Gardner, M.: Do nlp models know
  numbers? probing numeracy in embeddings. In: Proceedings of the 2019
  Conference on Empirical Methods in Natural Language Processing and the 9th
  International Joint Conference on Natural Language Processing (EMNLP-IJCNLP).
  pp. 5307--5315 (2019)

\bibitem{wang2019bert}
Wang, A., Cho, K.: {BERT} has a mouth, and it must speak: {BERT} as a markov
  random field language model. arXiv preprint arXiv:1902.04094  (2019)

\bibitem{wang2018stealing}
Wang, B., Gong, N.Z.: Stealing hyperparameters in machine learning. In: 2018
  IEEE Symposium on Security and Privacy (SP). pp. 36--52. IEEE (2018)

\bibitem{wolf2020transformers}
Wolf, T., Debut, L., Sanh, V., Chaumond, J., Delangue, C., Moi, A., Cistac, P.,
  Rault, T., Louf, R., Funtowicz, M., et~al.: Transformers: State-of-the-art
  natural language processing. In: Proceedings of the 2020 conference on
  empirical methods in natural language processing: system demonstrations. pp.
  38--45 (2020)

\bibitem{xu2014survey}
Xu, Y., Ma, T., Tang, M., Tian, W.: A survey of privacy preserving data
  publishing using generalization and suppression. Applied Mathematics \&
  Information Sciences  \textbf{8}(3), ~1103 (2014)

\bibitem{yousefpour2021opacus}
Yousefpour, A., Shilov, I., Sablayrolles, A., Testuggine, D., Prasad, K.,
  Malek, M., Nguyen, J., Ghosh, S., Bharadwaj, A., Zhao, J., et~al.: Opacus:
  User-friendly differential privacy library in pytorch. arXiv preprint
  arXiv:2109.12298  (2021)

\bibitem{yu2021differentially}
Yu, D., Naik, S., Backurs, A., Gopi, S., Inan, H.A., Kamath, G., Kulkarni, J.,
  Lee, Y.T., Manoel, A., Wutschitz, L., et~al.: Differentially private
  fine-tuning of language models. arXiv preprint arXiv:2110.06500  (2021)

\bibitem{zanella2020analyzing}
Zanella-Beguelin, S., Wutschitz, L., Tople, S., Ruhle, V., Paverd, A.,
  Ohrimenko, O., Kopf, B., Brockschmidt, M.: Analyzing information leakage of
  updates to natural language models. In: Proceedings of the 2020 ACM SIGSAC
  Conference on Computer and Communications Security. pp. 363--375 (2020)

\bibitem{zhang2022survey}
Zhang, H., Song, H., Li, S., Zhou, M., Song, D.: A survey of controllable text
  generation using transformer-based pre-trained language models. arXiv
  preprint arXiv:2201.05337  (2022)

\bibitem{zhang2018privacy}
Zhang, T., He, Z., Lee, R.B.: Privacy-preserving machine learning through data
  obfuscation. arXiv preprint arXiv:1807.01860  (2018)

\bibitem{zhu2020more}
Zhu, T., Ye, D., Wang, W., Zhou, W., Yu, P.: More than privacy: Applying
  differential privacy in key areas of artificial intelligence. IEEE
  Transactions on Knowledge and Data Engineering  (2020)

\end{thebibliography}

\appendix

\section{Extended Dataset Preprocessing}
\label{sec:extended_preprocessing}
The following sections contain additional details about the preprocessing. In particular, we  give further information about the preparation of the Enron dataset, and provide an overview of the label distributions.

\subsection{Preprocessing the Enron dataset}
As mentioned in Section~\ref{sec:datasets}, the Enron dataset does not come with categorical labels. The raw dataset contains the text of the email messages, alongside additional headers with meta-information. One of these headers called ``X-folder” contains the location of the email in folder structure of each user. We used the last folder name in the location paths to create labels. While these folders are unique to each user, some general overlap between the naming conventions and folder contents made it possible to select seven folders as labels that can be used for single-label text classification. These labels are: ``logistics'', ``personal'', ``management'', ``deal discrepancies'', ``resumes'', ``online trading'', and ``corporate''.

While preprocessing the text of the emails, we found that the message bodies contained a great amount of message chains, where some meta-information from the email headers (such as lists of email addresses, network information, message ID) were also present between the texts. 
This turned out to have a negative effect on output of our models when generating text. 
Namely, the generated samples lost all coherence and contained a lot of random concatenations of subword tokens. 
Therefore, we decided to discard all reply and forward chains from the emails by removing the parts following phrases that indicate a replied or forwarded message. This additional preprocessing, alongside the removal of HTML links, substantially improved the diversity and quality of the generated text samples.

\subsection{Label Distributions}
\begin{figure}
\centering
\begin{subfigure}{.5\textwidth}
  \centering
  \includegraphics[width=0.8\linewidth, height=4.5cm]{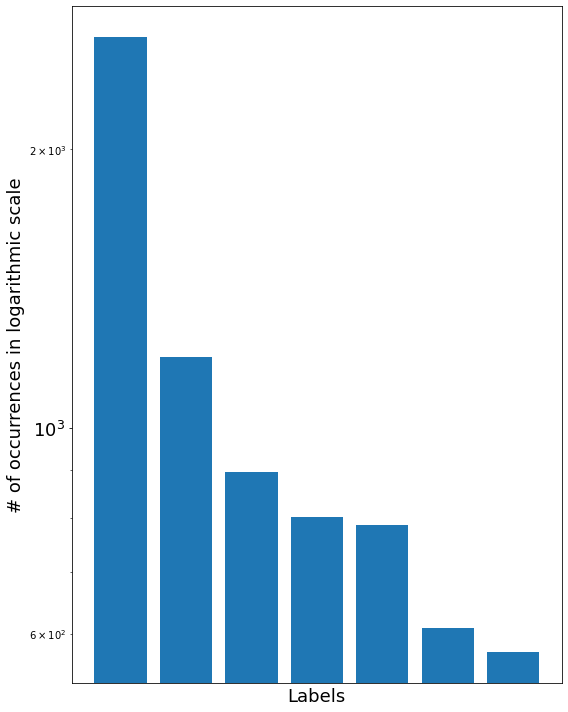}
  \caption{Enron}
  \label{fig:label_enron}
\end{subfigure}%
\begin{subfigure}{.5\textwidth}
  \centering
  \includegraphics[width=0.8\linewidth, height=4.5cm]{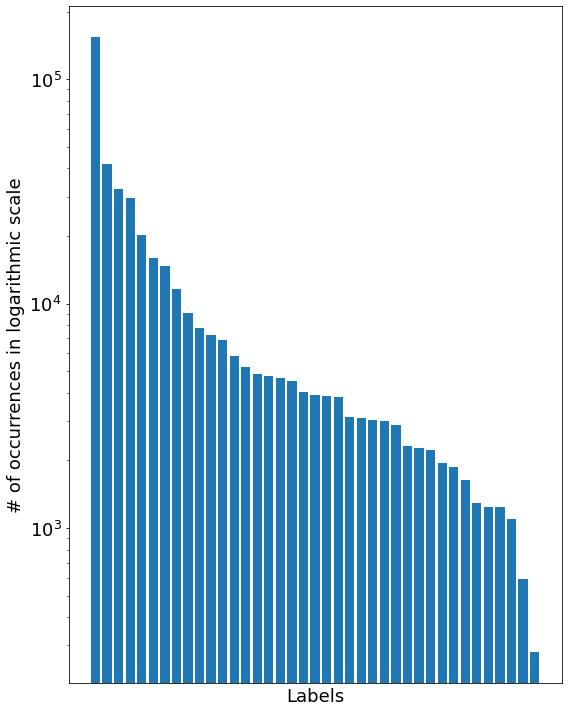}
  \caption{Blog Authorship}
  \label{fig:label_blogauth}
\end{subfigure}
\caption{Label distribution of the datasets}
\label{fig:test}
\end{figure}

Figures~\ref{fig:label_enron} and~\ref{fig:label_blogauth} show the label distribution of both datasets. 
In the Enron dataset, the most frequent label is ``personal” with 2,062 occurrences, while on the Blog Authorship dataset the label ``Student” has the highest count with 153,903.  
During the train-test split, the distribution of the dataset was retained in both splits.

\section{Extended Experiment Results}
\label{sec:extended_experiments}
This section contains the extended experimental results. We provide an extended review of the NER results with detailed description of the entity types, a discussion about selecting specific entity types for our study, and the additional named entity extraction results not present in the main paper.

\subsection{Named Entity Recognition}
\label{sec:extended_ner}
As mentioned in Section~\ref{sec:procedure_ner}, we used the spaCy  library~\cite{zenodo121230} for the process of NER.
The named entity recognizer of spaCy distinguishes between the following 18 entity types:

\begin{itemize}
    \item \textbf{PERSON} - people names, including fictional
    \item \textbf{NORP} - nationalities or religious or political groups
    \item \textbf{FAC} - facilities - building, airports, bridges, etc.
    \item \textbf{ORG} - organizations, companies, agencies, institutions
    \item \textbf{GPE} - geopolitical entities - countries, cities, states
    \item \textbf{LOC} - non-GPE locations
    \item \textbf{PRODUCT} - objects, vehicles, foods
    \item \textbf{EVENT} - named hurricanes, wars, sport, events etc
    \item \textbf{WORK\_OF\_ART} - titles of books, songs, etc
    \item \textbf{LAW} - named documents made into laws
    \item \textbf{LANGUAGE} - any named language
    \item \textbf{DATE} - absolute or relative dates, periods
    \item \textbf{TIME} - times smaller than a day
    \item \textbf{PERCENT} - percentages
    \item \textbf{MONEY} - monetary values, including unit
    \item \textbf{QUANTITY} - Measurements, as of weight or distance
    \item \textbf{ORDINAL} - ``first'', ``second'', etc
    \item \textbf{CARDINAL} - numerical values not covered by other types
\end{itemize} 

The initial results of the NER can be seen in Table~\ref{tab:ext_named_ents}. Compared to the values in Table~\ref{tab:named_ents}, these counts also include the repeated occurrences of a named entity.

In studying the extent of named entity memorization, we focused on entity types that have a higher probability to contain personal or privacy sensitive information. \textit{PERSON} can include first and last names, which together can be considered as personal information. \textit{ORG}, \textit{GPE}, \textit{LOC} and \textit{FAC} include information that can be pieced together to identify a likely data subject, therefore they also fit into the personal information category. We included \textit{Money}, as in some industries (\ie banking) records of specific amounts can be regarded as privacy sensitive information. Finally, we also included \textit{CARDINAL} as this type can refer to card numbers, phone numbers, and different ID numbers, which can be both personal and privacy sensitive.

\begin{table}[ht]
    \small
    \centering
    \caption[Extended Named entity recognition results]{Per type number of named entities in the datasets}\label{tab:ext_named_ents}
    \begin{tabular}{lrr}
    \toprule
    Named Entity Type & Enron & Blog Authorship\\  
    \midrule                                                       
    \hline
    PERSON       & 33,993 & 767,144\\ 
    NORP         & 1,373 & 132,080\\
    FAC          & 674 & 24,405\\
    ORG          & 30,365 & 521,159 \\
    GPE          & 8,771 & 298,607\\
    LOC          & 827 & 37,286\\
    PRODUCT      & 992 & 27,432\\
    EVENT        & 333 & 14,131\\
    WORK\_OF\_ART & 1,354 & 43,102\\
    LAW          & 151 & 5,798\\
    LANGUAGE     & 69 & 11,161\\
    DATE         & 17,319 & 626,969\\
    TIME         & 7,137 & 211,993\\
    PERCENT      & 579 & 17,332\\
    MONEY        & 2,952 & 39,768\\
    QUANTITY     & 655 & 21,512\\
    ORDINAL      & 1,210 & 104,559\\
    CARDINAL     & 17,725 & 444,764\\
    \bottomrule
    \end{tabular}
\end{table}

\subsection{Text Generation}
\label{app:text_generation}
This section includes text samples generated from our fine-tuned models. 
Tables~\ref{tab:full_samples}, \ref{tab:partial_samples}, and~\ref{tab:dp_full_samples} show randomly selected sequences of generated text under the \textit{Full}, \textit{Partial}, and \textit{Differentially Private} fine-tuning setting for each of the two datasets and prompting strategies.

\begin{table*}[ht]
    \scriptsize
    \centering
    \caption{Randomly selected sequences of generated text from the \textit{Full} setup}\label{tab:full_samples}
    \begin{tabular}{l|r|r}
    \toprule
      Generated Sample & Dataset & Prompt method\\
    \midrule
      \multicolumn{1}{p{10cm}|}{\raggedright Organizations from among 150 organizations and national educational leaders Global Quality Internet Service Vision Vision Watch Associates Association National / CT Science Vision Health network} & Enron & naive\\
      \hline
    \multicolumn{1}{p{10cm}|}{\raggedright In, Justice \& Justice : International / Research Papers Issues to 2017. : International Organizations Research in Business Fellow and Masiah Al Hehar and Professor and Chair, University and Karachi Emeritus, from Durham } & Enron & informed\\
     \hline
    \multicolumn{1}{p{10cm}|}{\raggedright Third to the Joint Chiefs, between 2015 - 2020 side directors permanent appointment playing two leaders joint in training urban core areas in economic and medical development organisations in electoral the 2020 European Parliament} & Blog Authorship & naive\\
    \hline
    \multicolumn{1}{p{10cm}|}{\raggedright Sheriff Avery and Judge said everyone and his girls were safe and they all was riot! Only Judge White and the churchmaster had not heard tomorrow, and things was right Maydal Sio and all his lawyers were punishment}         & Blog Authorship & informed \\
    \bottomrule
    \end{tabular}
\end{table*}

\begin{table*}[ht]
    \scriptsize
    \centering
    \caption{Randomly selected sequences of generated text from the \textit{Partial} setup}\label{tab:partial_samples}
    \begin{tabular}{l|r|r}
    \toprule
      Generated Sample & Dataset & Prompt method\\
    \midrule
      \multicolumn{1}{p{10cm}|}{\raggedright Annex 1 at Anne. M. 1, featuring new entrances and entrance towards the entrances from East Village and West Village Parkway ( connecting to and from both tunnels into and from West Village ) also includes the entire structure encompassing the new tunnels on all four floors including the main floors plus all 4 rooms} & Enron & naive\\
      \hline
    \multicolumn{1}{p{10cm}|}{\raggedright Sir Christopher and to Commander Fox, presented a statement that they owed to Saunders for saving Christopher before sailing for Britain. Saunders blamed Christopher ; they were wrong, obviously not, and clearly angered they were, but Adams insisted Christopher instead was " damaged, was very " sick and " emotionally "}         & Enron & informed\\
     \hline
    \multicolumn{1}{p{10cm}|}{\raggedright Those last day. Those those strange voices - Those eerie voices - That haunted voices - that haunting voice, That haunting ghost - that echoing in echoes of the haunting? Why silence? Why? - The ghost? Which is it? who is ghost And silent voice?} & Blog Authorship & naive\\
    \hline
    \multicolumn{1}{p{10cm}|}{\raggedright  Paul. Paul leaves away before that scene breaks as Helen laughs bitterly with Julia and Liz. Paul hates love and they hate him because they love him so bad so love, her toils against Paul and hasces with Paul. Paul makes love Simon}         & Blog Authorship & informed \\
    \bottomrule
    \end{tabular}
\end{table*}

\begin{table*}[ht]
    \scriptsize
    \centering
    \caption{Randomly selected sequences of generated text from the \textit{Differentially Private} setup}\label{tab:dp_full_samples}
    \begin{tabular}{l|r|r}
    \toprule
      Generated Sample & Dataset & Prompt method\\
    \midrule
      \multicolumn{1}{p{10cm}|}{\raggedright exampleenerMFuzkt few1 thateg5 talkden landkin peopledan mentionilialiltendan1tagB landseB ofsemadlan mouth talkrodrodelinetieei mouth mouthmadpenukensorlattensorelepedlen start1ndimilebuuzseuzalaeleicidanici mouthisipen } & Enron & naive\\
      \hline
    \multicolumn{1}{p{10cm}|}{\raggedright mouthsixhiluerikntonhillattlattedelenacherwearelinestlubchorileelinsionenerpenc ionpenkineonelinvilsettsablepencardtinxinpensablelansorNmileelinbenkinkinndiglielinkinxinpenpenelinchinelipeth"}         & Enron & informed\\
     \hline
    \multicolumn{1}{p{10cm}|}{\raggedright ,. " in him not guitar around them no brother was time and no The up lap up wasly guitar no againul un from without lap aside " sex of time as " The if them a no? " aside all around aside around un.. down t O un from guitar aside without without O O him aside " No A " in time brother guitarist unul The The in,, him was out with. } & Blog Authorship & naive\\
    \hline
    \multicolumn{1}{p{10cm}|}{\raggedright lap in No the The while roundul guitar no was the guitarcase without was aside the " without A without t guitars " round while, o team as like draw O O as done small o up not was and and out out away in in. the aside without guitarist., until t the team away. at down no not up down " if t up the guitar as of the. in away, un away }         & Blog Authorship & informed \\
    \bottomrule
    \end{tabular}
\end{table*}

\end{document}